\documentclass[letterpaper, 10 pt, conference]{ieeeconf}  
\IEEEoverridecommandlockouts                              

\usepackage{graphicx}
\graphicspath{{figures/}
{/a/second/directory/path/can/go/here/}}

\usepackage{caption}
\usepackage{subcaption}

\usepackage{algorithm}

\usepackage[noend]{algpseudocode}
\usepackage{setspace}

\usepackage{hyperref}
\usepackage{bm}
\usepackage[load-configurations=version-1]{siunitx} 

\usepackage{xcolor}
\usepackage{balance}
\usepackage{amsmath, amssymb}
\DeclareMathOperator{\dist}{d}
\DeclareMathOperator*{\argmax}{arg\,max}

\title{\LARGE \bf
Hierarchical Reinforcement Learning for Quadruped Locomotion
}

\author{Deepali Jain, Atil Iscen, and Ken Caluwaerts
\thanks{Robotics at Google, 10011 New York, USA.}
\thanks{\{jaindeepali, atil, kencaluwaerts\}@google.com}
}

\begin{document}

\maketitle
\thispagestyle{empty}
\pagestyle{empty}

\begin{abstract}

Legged locomotion is a challenging task for learning algorithms, especially when the task requires a diverse set of primitive behaviors. 
To solve these problems, we introduce a hierarchical framework to automatically decompose complex locomotion tasks. 
A high-level policy issues commands in a latent space and also selects for how long the low-level policy will execute the latent command. 
Concurrently, the low-level policy uses the latent command and only the robot's on-board sensors to control the robot's actuators.
Our approach allows the high-level policy to run at a lower frequency than the low-level one. 
We test our framework on a path-following task for a dynamic quadruped robot and 
we show that steering behaviors automatically emerge in the latent command space as low-level skills are needed for this task. 
We then show efficient adaptation of the trained policy to a different task by transfer of the trained low-level policy. Finally, we validate the policies on a real quadruped robot.
To the best of our knowledge, this is the first application of end-to-end hierarchical learning to a real robotic locomotion task.

\end{abstract}

\section{INTRODUCTION}

Locomotion for legged robots is a challenging control problem that requires high-speed  control of actuators as well as precise coordination between multiple legs based on various types of sensor data. In addition to basic locomotion, different terrains, tasks or environmental conditions might require specific primitive behaviors. 

Recent research shows promising results on learning based systems for locomotion tasks in simulation and real hardware~\cite{hwangbo2019learning,iscen2018policies,yu2018policy}. 
Various techniques can be used to discover policies for such tasks. In this work, we focus on Reinforcement Learning (RL) to obtain robust policies.

Robot locomotion is an excellent match for hierarchical control architectures. Indeed, the separation of low-level control of the legs and high-level decision making based on the environment and task at hand provides multiple advantages such as reuse of the learned low-level skills across tasks, and interpretability of the high-level decisions.

Given a complex task, manually defining a suitable hierarchy is typically a tedious task that requires engineering of the state and action spaces as well as reward functions for each primitive. To overcome this, we introduce a hierarchical framework to automatically decompose complex locomotion tasks. A high-level policy issues commands to a low-level policy and decides for how long to execute the low-level policy at a time. The low-level policy acts according to commands from the high-level policy and on-board sensors. Our approach allows separation of the state variables that are used for low-level control, from state variables only required for higher-level control. Our architecture naturally allows the high-level to operate at a slower timescale than the low-level.

We test our framework on a path following task for a dynamic quadruped robot. The task requires walking into different directions to complete the track while keeping balance. Using our architecture, we train both levels of the hierarchical policy end-to-end. We show that  steering behavior automatically emerges in the latent command space between the high-level and low-level policies, which allows  reuse of the learned low-level behaviors. We show  transfer of the low-level policy to a different track to achieve fast adaptation to a new task. Lastly, we deploy our policies to hardware to validate the learned behaviors on a real robot.

\begin{figure}[tbp]
\centering
\includegraphics[width=0.9\linewidth]{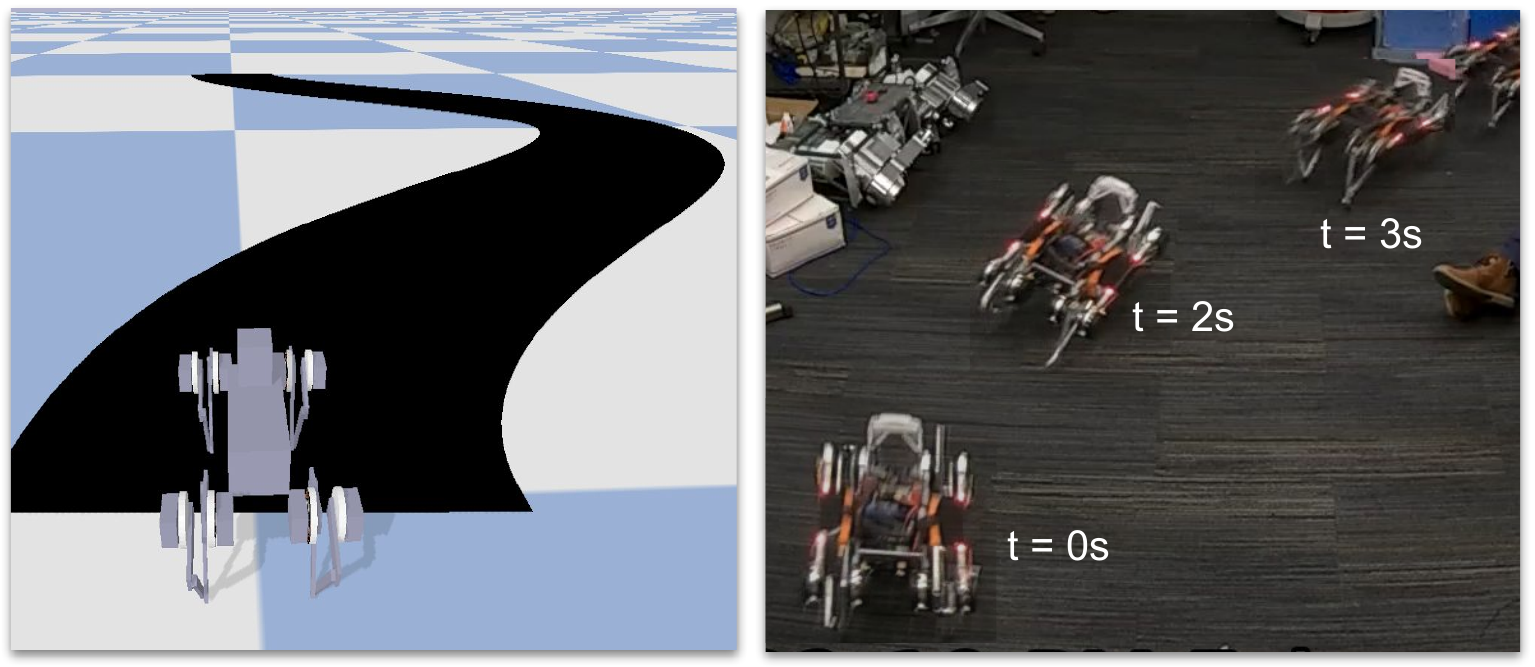}
\caption{Simulated task on the left and the robot performing a hierarchical policy learned in simulation. During execution the high-level policy executes intermittently to update the latent command for the low-level policy.}
\label{fig:intro-path}
\end{figure}





\section{RELATED WORK}
Hierarchical Reinforcement Learning (HRL) methods focus on decomposing complex tasks into simpler sub-tasks. Not only does this help simplify a single difficult problem, it can also help in adapting the solution faster to a new problem if sub-tasks are general enough. The framework based on pre-defined options~\cite{sutton1999between}, or temporally extended actions, is one of the first popular methods in this direction. More recently, considerable research attention is given to the problem of automatically discovering options through experience. 

In methods like HRL with hindsight~\cite{levy2018hierarchical} and data-efficient HRL~\cite{nachum2018data}, hierarchy is introduced using universal value functions (value functions that are parameterized by 'goal'). Actions of a higher-level policy, running at a fixed slower timescale, act as goals for a lower-level. A goal is explicitly defined as a point in observation space and the low-level is rewarded for reaching that point. This allows both levels to be trained through their respective reward signals. However, this goal specification is not suitable in all situations. If the observation space is high dimensional, then the high-level task of selecting a goal becomes very difficult. Also, determining when the goal is achieved requires task-specific domain knowledge.

Latent space policies for HRL~\cite{haarnoja2018latent} use a different approach to parameterize the low-level. The high-level outputs a set of latent variables as goal for the lower level that are learned through maximum entropy reinforcement learning. Both levels are then trained to maximize the main task reward. This, however, prevents the low-level from being reused for any other task. 

Along similar lines, Osa et. al.~\cite{osa2019hierarchical} recently proposed a method based on information maximization to learn latent variables of a hierarchical policy. 

In their paper on meta learning shared hierarchies~\cite{frans2017meta}, Kevin et al. propose a HRL framework that is learned on multiple related tasks. The low-level skills are reused across tasks while the meta-controller is task-specific. Instead of parameterizing a single low-level policy, the meta-controller selects a different low level policy from a set for each sub-task. In order for general low-level policies to emerge, the framework needs to be trained on a number of related tasks.

In our method, we use a latent goal representation to remove the need to hand design low-level rewards or deciding on the number of low-level policies. We also use different state representations for both levels to ensure that reusable low-level skills are learned even when trained on a single task. Moreover, in our method, the high-level policy runs at a variable timescale, easing processing requirements for higher-level state information. 

The task of robot navigation lends itself to a hierarchical solution with path-planning at the high-level and point-to-point locomotion at the low-level. In this context, many methods~\cite{bischoff2013hierarchical, heess2016learning, faust2018prm} have been tried to solve these two tasks separately. Nicolas et al.~\cite{heess2016learning}, propose a hierarchical framework for locomotion based on modulated locomotor controllers. A low-level \emph{spinal} network learns primitive locomotion by training on simple tasks. A high-level \emph{cortical} network, drives behavior by modulating the inputs to the pre-trained spinal network. HRL with pre-trained primitives is also applied to the task of robot locomotion on rough terrains~\cite{peng2017deeploco, peng2016terrain}. In the DeepLoco~\cite{peng2017deeploco} paper, low-level controllers achieve robust walking gaits that satisfy a stepping-target. High-level controllers then invoke desired step targets for the low-level controller. 

We apply our hierarchical learning method to the robot locomotion task of following a path in 2D. Our method does not need specification of timescales for the two levels nor a low-level reward signal. Our end-to-end hierarchical learning framework automatically discovers steering behaviors at the low-level which can transfer to a real quadruped robot.

\section{METHOD}

\subsection{Hierarchical Policy Structure and Execution}
\begin{figure}[hbpt]
\centering
\includegraphics[width=0.8\linewidth]{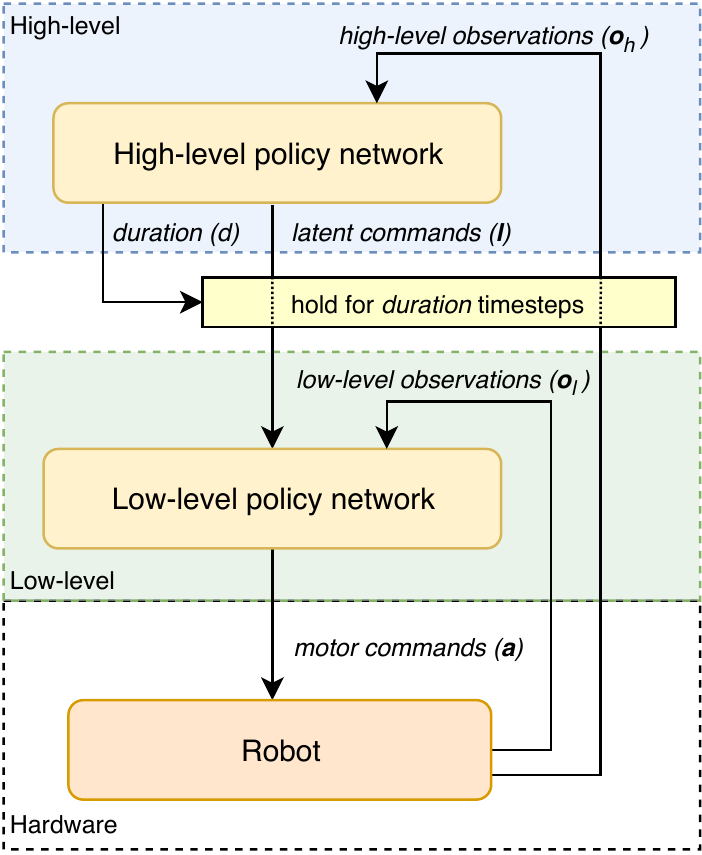}
\caption{Hierarchical policy. The high-level policy with parameters $\bm{\phi}_h$ receives high-level observations $\bm{o}_h$ and outputs a \emph{latent command vector} $\bm{l}$ and a duration $d$. The low-level policy (parameters  $\bm{\phi}_l$) computes motor commands $\bm{a}$ based on  $\bm{l}$ and low-level observations $\bm{o}_l$. The high-level policy is only evaluated every $d$ steps. The architecture is trained end-to-end.}
\label{fig:hrlPolicy1}
\end{figure}

\begin{figure*}
\centering
\includegraphics[width=0.8\linewidth]{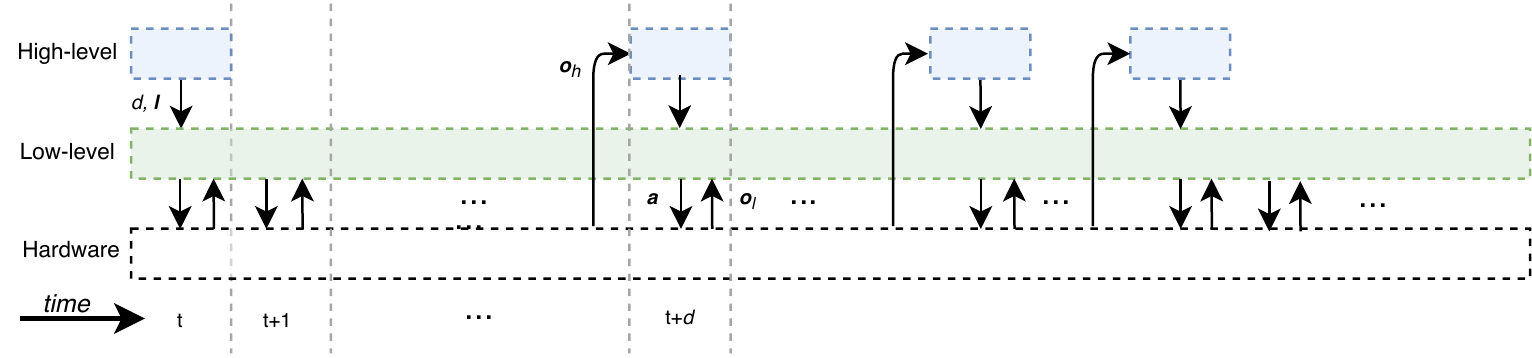}
\caption{Hierarchical policy evaluation timeline. The high-level policy computes a latent command for the low-level policy and a duration for which to execute the low-level policy. The low-level policy interacts with the hardware at a constant frequency. At the end of the high-level period, the high-level receives updated high-level observations and computes a new latent command and duration.}
\label{fig:policy_duration}
\end{figure*}

Our hierarchical policy is structured as shown in Fig.~\ref{fig:hrlPolicy1}. The high-level policy (HL) receives higher-level observations from the environment  and issues commands in a latent space to a low-level policy. The high-level also decides the duration for which the low-level is executed before the next high-level evaluation. 
The low-level (LL) receives observations from on-board sensors (low-level) and the current latent command from the high-level. 
It outputs actions to execute on the hardware. At the end of the duration set by the high-level, the high-level is invoked again and the process repeats (Fig.~\ref{fig:policy_duration}). 
Both high-level and low-level policies in this architecture are neural networks. Algorithm~\ref{alg:rollout} shows how an episode is executed using a hierarchical policy in which the high-level and low-level have weights $\bm{\phi}_h$ and $\bm{\phi}_l$ respectively.  

\begin{algorithm}
\caption{Executing a Hierarchical Policy}\label{alg:rollout}
\begin{algorithmic}[1]
\Procedure{RunHRL}{$\bm{\theta}$}\Comment{\parbox[t]{.4\linewidth}{HRL policy weights}}
 \State \{$\bm{\phi}_h, \bm{\phi}_l \} = \bm{\theta}$
 \State $\bm{o}_h \gets$ initial HL observation
 \State $R \gets 0$\Comment{\parbox[t]{.4\linewidth}{Episode reward}}
 \State $d \gets 0$ \Comment{\parbox[t]{.4\linewidth}{LL duration}}
 \While{not end of episode}
  \If{$d = 0$}
    \State $\bm{o}_h \gets$ HL observation
    \State \{$d, \bm{l}\} \gets f_{\bm{\phi}_h}(\bm{o}_h)$ \Comment{\parbox[t]{.4\linewidth}{Duration, latent command}}
  \EndIf
  \State $\bm{a} = f_{\bm{\phi}_l}(\bm{o}_l, \bm{l})$ \Comment{\parbox[t]{.5\linewidth}{\mbox{LL action (motor commands)}}}
  \State $\bm{o}_l, r \gets$ StepInEnvironment($\bm{a}$)
  \State $d \gets d -1$
  \State $R \gets R + r$
 \EndWhile
\State \textbf{return} $R$\Comment{\parbox[t]{.5\linewidth}{Total reward for the episode}}
\EndProcedure
\end{algorithmic}
\end{algorithm}

\subsection{Learning Parameters of a Hierarchical Policy}
To jointly learn the parameters of the high-level and low-level neural networks, we optimize a standard reinforcement learning objective. Consider a state space $\mathcal{S}$ and action space $\mathcal{A}$. A sequential decision making or control problem can be modeled as a Markov Decision Process (MDP). An MDP is defined by a transition function $P(\bm{s}_{t+1} | \bm{s}_t, \bm{a}_t)$ and a reward function, $r(\bm{s}_t, \bm{a}_t)$. A policy $\pi_\mathbf{\theta}(\bm{s})$, parameterized by a weight vector $\bm{\theta}$, maps states $\bm{s}$ to actions $\bm{a}$. For a hierarchical policy, $\bm{\theta}$ is the collection of parameters from all levels ($\bm{\theta}=\{\bm{\phi}_h, \bm{\phi}_l\}$) and the subset of state variables observable by the high-level and low-level are denoted as $\bm{o}_h$ and $\bm{o}_l$ respectively. The policy interacts with the MDP for an episode of $T$ timesteps at a time. The reinforcement learning objective is to maximize the expected total reward at the end of episode:
\begin{equation}
    \argmax_\theta \quad \mathbb{E}\left[\sum_{t=1}^Tr\left(\bm{s}_t, \pi_\mathbf{\theta}\left(\bm{s}_t\right)\right)\right].
\end{equation}

We use a simple derivative-free optimization algorithm called Augmented Random Search (ARS)~\cite{mania2018simple} to maximize $R$. The algorithm proceeds by choosing a number of directions uniformly at random on a sphere in policy parameter space, then evaluates the policy along these directions and finally updates the parameters along the top performing directions.


\subsection{Transferring Low-Level Policies}
An interesting aspect of our hierarchical method is that after learning a policy on one task, the low-level policy can be transferred to a new  task from a similar domain. This allows sharing of primitive skills across related problems and is faster than learning from scratch on each task. The low-level policy can be transferred by keeping $\bm{\phi}_l$ fixed after learning on the original task and re-initializing $\bm{\phi}_h$. Then, during training only $\bm{\phi}_h$ is updated by ARS.



\section{EXPERIMENTS}
\subsection{Task Details}
We apply our method to a path-following  task for a quadruped robot. For this, we use the Minitaur quadruped robot from Ghost Robotics\footnote{\href{https://www.ghostrobotics.io/robots}{ghostrobotics.io}}. The Minitaur robot has $8$ degrees of freedom ($2$ per leg). The swing and extension of each the legs is controlled using a PD position controller provided with the robot. We train our policies in simulation using pyBullet~\cite{pybulletcoumans,tan2018sim}.

For the locomotion task, we tackle the problem of following a curved path in 2D while staying within the allowed region. The robot is rewarded for moving towards the end of the path. The task requires the robot to steer left and right at different angles. The optimal trajectory for the center of mass for the robot is not defined and  depends on the robot's anatomy and learned low-level behaviors. 
Steering poses additional challenges because the legs of the robot can only move in the sagittal plane. 
The reward function is given by:
\begin{align}
    r(t) &= \dist\left(\bm{x}(t-1), \bm{x}^{\mbox{goal}}\right) - \dist\left(\bm{x}(t), \bm{x}^{\mbox{goal}}\right)\\
    R &= \sum_{t \ge 1}r(t),
\end{align}
where $\dist(., .)$ is the Euclidean distance, $\bm{x}$ is the position of the robot, and $\bm{x}^{\mbox{goal}}$ is the final position of the path. 
We terminate an episode as soon as the robot moves out of the path.

To learn locomotion, we use the recent \emph{Policies Modulating Trajectory Generators} (PMTG) architecture,  which has shown success at learning forward locomotion on quadruped robots~\cite{iscen2018policies}. The PMTG architecture takes advantage of the cyclic characteristic of locomotion and of leg movement primitives by using trajectory generators. Trajectory generators serve as parameterized functions that provide circular leg positions. The policy is responsible to modulate the generator and adjust leg trajectories with a residual as needed. A more detailed explanation of the architecture can be found in the paper~\cite{iscen2018policies}. Our hierarchical policy is responsible for controlling the PMTG architecture which issues motor position commands. 

\subsection{Hierarchical Architecture}
As demonstrated in previous work~\cite{iscen2018policies}, a well-trained linear neural network policy in combination with the PMTG can produce locomotion. Therefore we use linear neural networks for the high-level and the low-level policies. However, we clip the latent command space to $[-1, 1]^{\dim(\bm{l})}$, which allows us to more easily study the latent space. The number of dimensions of the latent command $\dim(\bm{l})$ is a hyper-parameter. 
Note that while the policy networks are linear, PMTG introduces recurrency and non-linearities~\cite{iscen2018policies}.

We separate the state information into two. We only feed the robot's position $\bm{x}$ and the robot's orientation (yaw direction) into the high-level policy (4-dimensional). The high-level policy outputs the latent command $\bm{l}$ and a \mbox{duration $d$}. 

The low-level policy network observes the 8-dimensional PMTG state (we use 4 trajectory generators, one per leg), 4-dimensional IMU sensor data (roll, pitch, roll rate, pitch rate), and the latent command $\bm{l}$ from the high-level policy. The output of the low-level network are $8$ motor positions and $8$ PMTG parameters. 

We update the low-level's output every $6\si{\milli\second}$. The high-level is executed every $d$ low-level steps (where $d$ was calculated during the previous high-level cycle).
In practice $d$ is rescaled to $[100, 700]$ from the $[-1, 1]$ clipped value. Since the low-level timestep is $6\si{\milli\second}$, the time between high-level evaluations is between $0.6\si{\second}$ and $4.2\si{\second}$. This highly simplifies the process of estimating the position and direction of the robot.

\subsection{Transfer of Low-Level Policies to New Tasks}
\begin{figure}
\centering
\begin{subfigure}{.5\textwidth}
  \includegraphics[width=.312\linewidth]{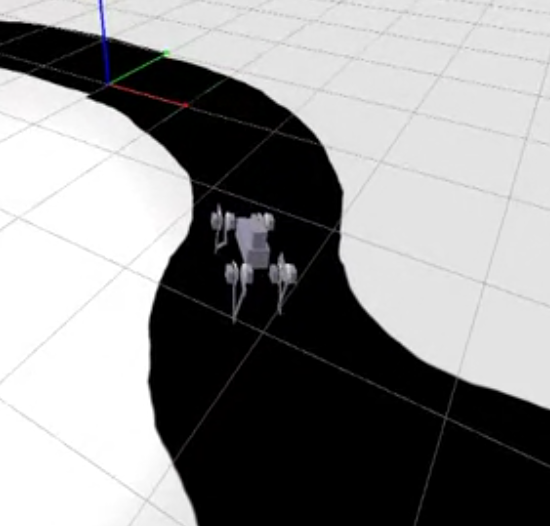}%
  \includegraphics[width=.325\linewidth]{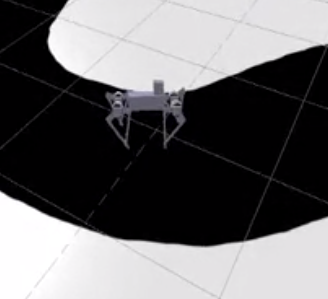}%
  \includegraphics[width=.312\linewidth]{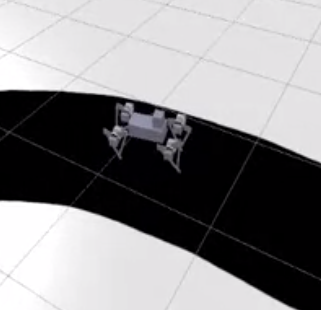}
  \caption{Robot path tracking in simulation. If the robot's center of mass exits the black area, the episode is terminated.}
  \label{fig:sim_paths_a}
\end{subfigure}

\begin{subfigure}{.5\textwidth}
  \includegraphics[height=.32\linewidth]{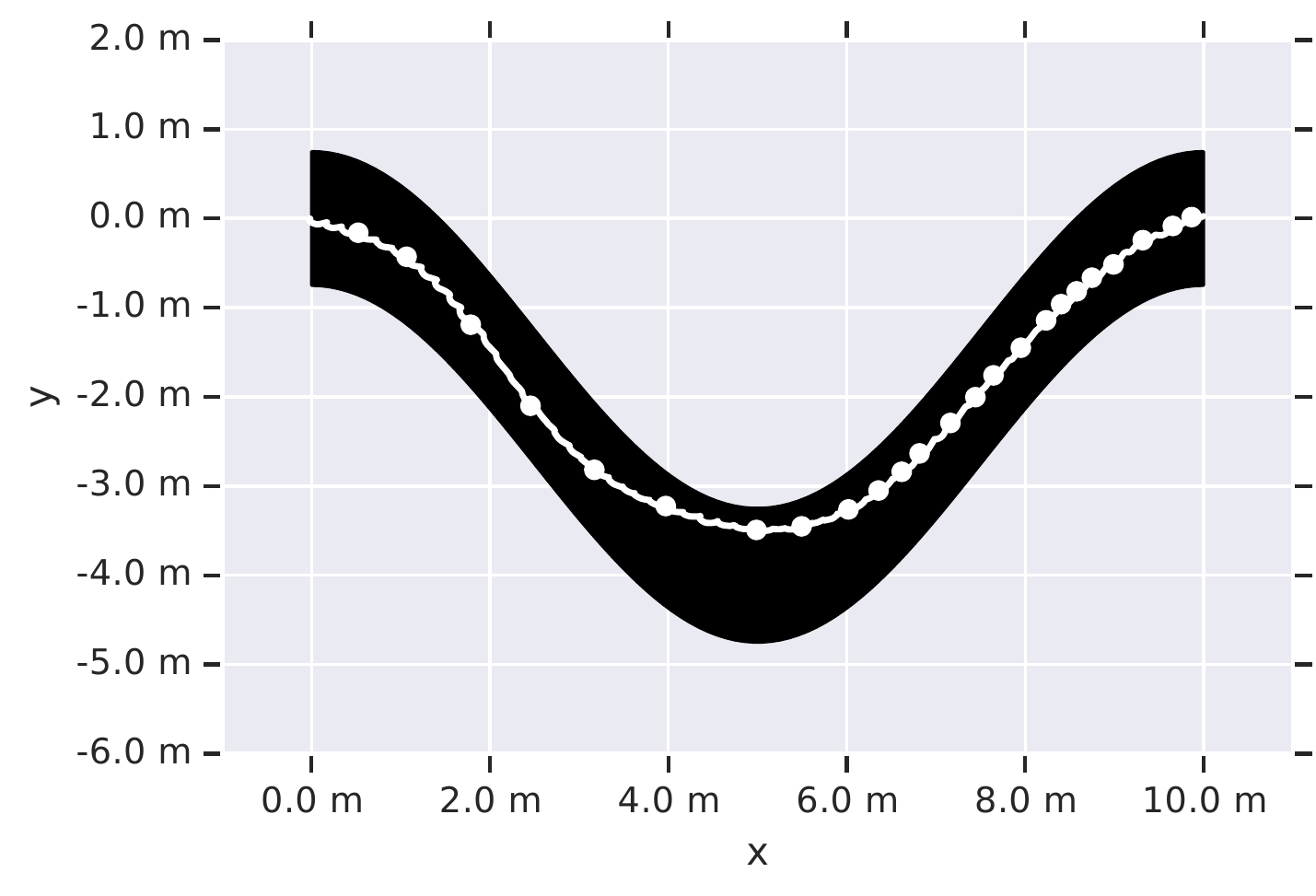}%
  \includegraphics[height=.32\linewidth]{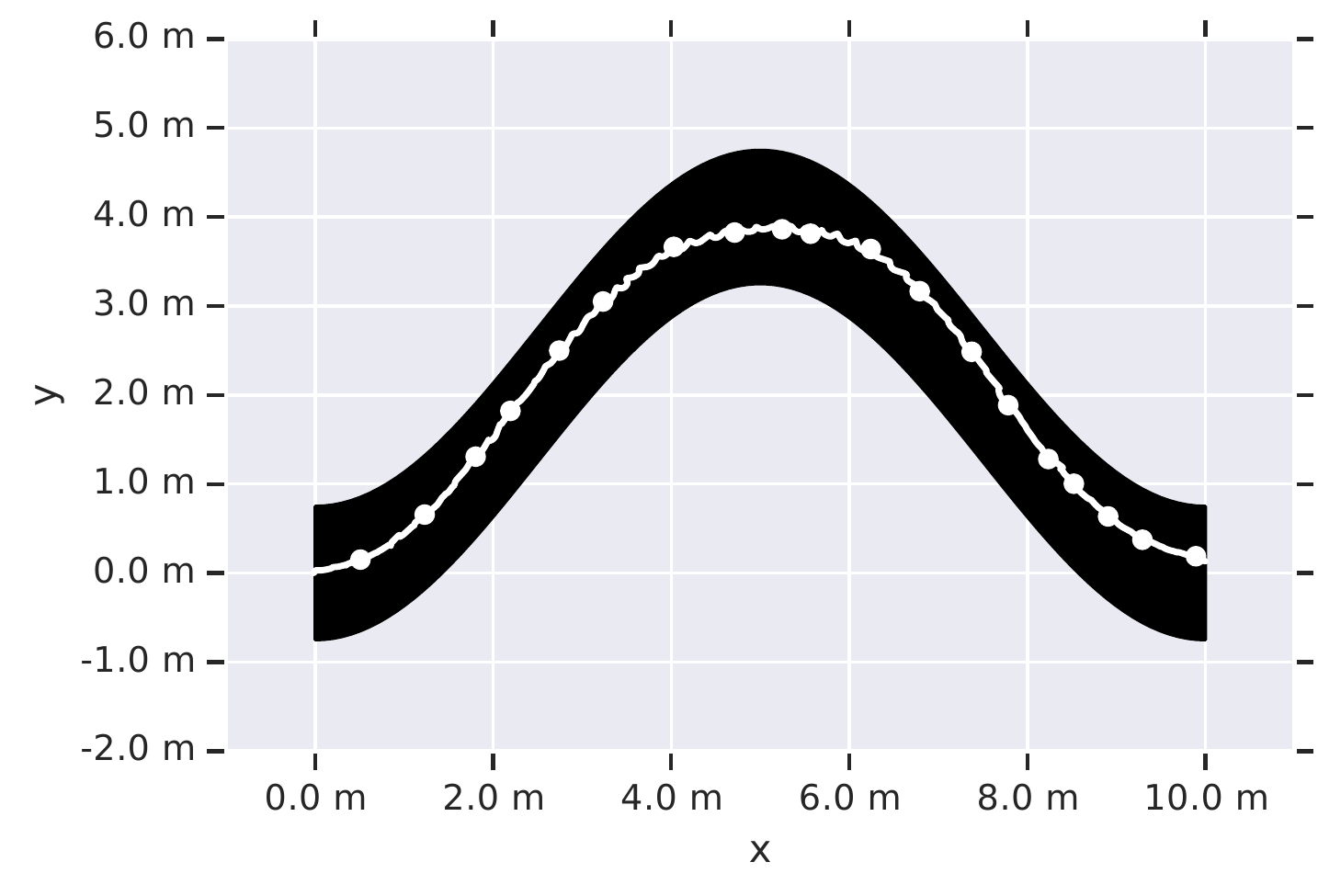}
  \caption{Trajectory on $2$ paths with a shared low-level policy (trained on the path on the left). Dots indicate when the high-level policy takes a new decision.}
  \label{fig:sim_paths_b}
\end{subfigure}
\caption{Sample rollouts in simulation of the path-tracking task with a 4D latent command space.}
\label{fig:sim_paths}
\end{figure}

We show that our architecture can adapt to $2$ different paths shown in Figure~\ref{fig:sim_paths}. We first train the architecture for path on the left side of Figure~\ref{fig:sim_paths}. The low-level policy only has access to proprioceptive sensor data and this forces it to learn  generic steering primitives that can be reused across different paths. We test this property of our hierarchical architecture by reusing the trained low-level policies from path $1$ when training on path $2$.

\subsection{Baselines}
For comparison, we train flat policies on these tasks. The input to the flat policies is the same as the high-level's observations concatenated with the low-level's in the hierarchical setup  (except, trivially, for the latent commands) and the output is the same as the low-level actions. The flat policy also uses the same PMTG architecture for a fair comparison.

Secondly, we implement an expert hierarchical policy for additional comparison. We pre-train the low-level policy for this baseline using a carefully designed and tuned reward function to follow a target steering angle. The high-level policy computes the running duration $d$ for the pre-trained low-level policy and also outputs a steering angle (a scalar in the range $-1$ (far left) to $1$ (far right), instead of the latent command $\bm{l}$). The input for the expert policy's high-level and low-level is exactly the same as in the HRL case.

As in the HRL case, the baseline policies are trained by directly optimizing $R$ using Augmented Random Search (ARS)~\cite{mania2018simple}. We perform evaluation across different search directions in parallel. We train each method with a set of hyper-parameters (number of directions to search in ARS, number of top direction for updating parameters and number of latent command dimensions in case of our hierarchical method). Finally, we pick the best hyper-parameter for each and compare the average performance of $5$ random training runs with those hyper-parameter settings.

\begin{figure}
\centering
\begin{subfigure}{.5\textwidth}
  \centering
  \includegraphics[width=.83\linewidth]{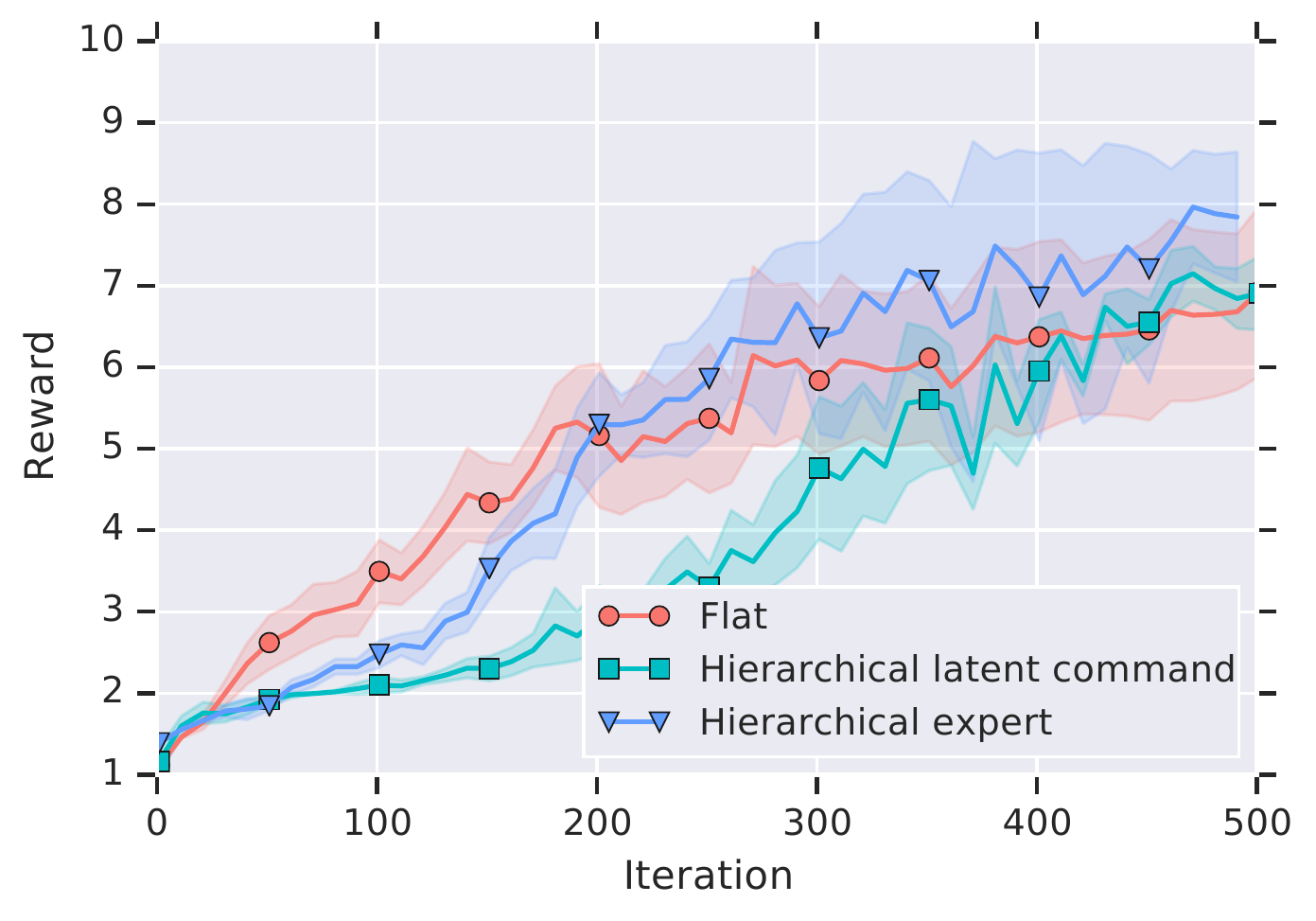}
  \caption{Learning curves for path 1. All policies are trained from scratch.}
  \label{fig:learning_curve_b}
\end{subfigure}
\begin{subfigure}{.5\textwidth}
  \centering
  \includegraphics[width=.83\linewidth]{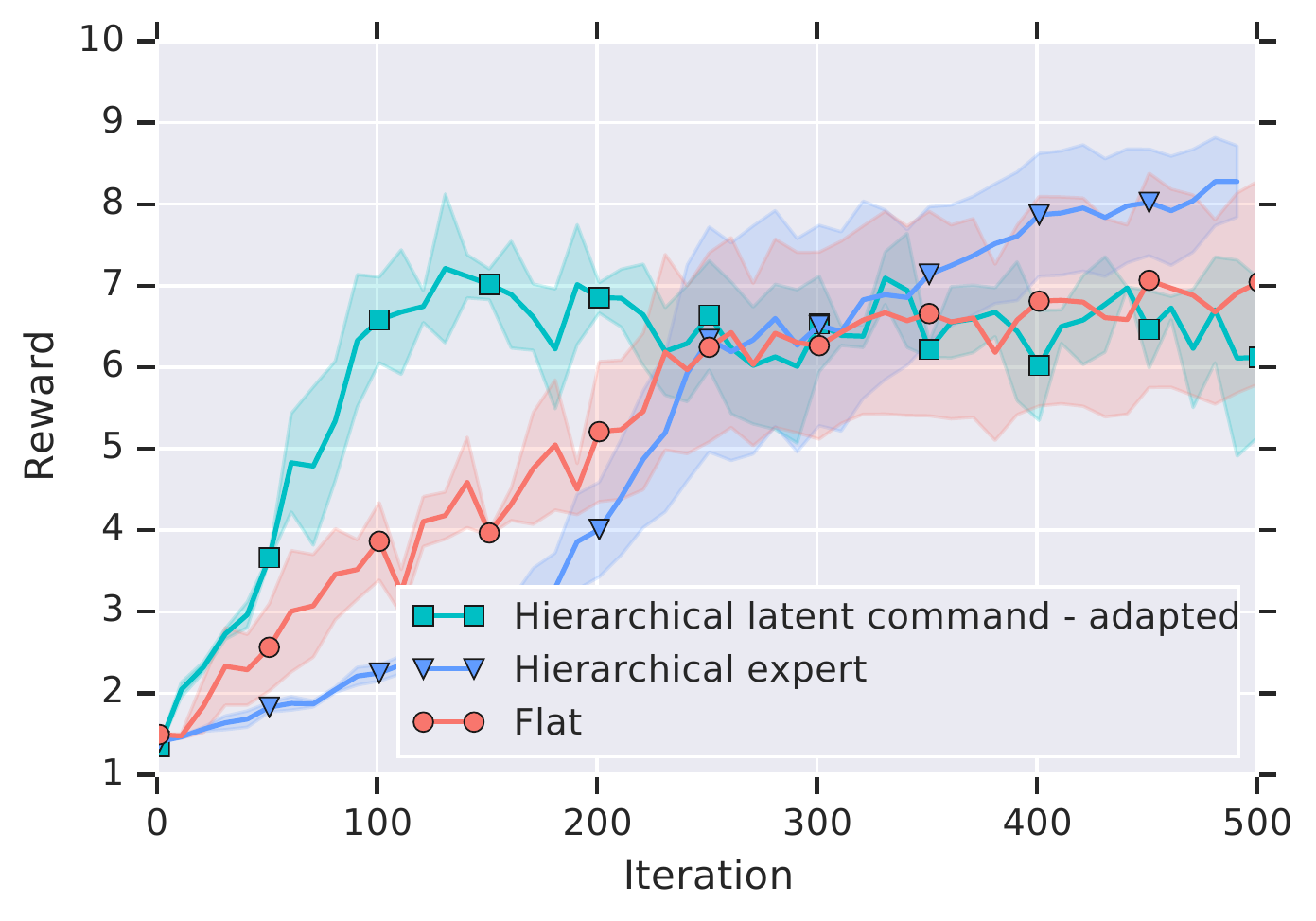}
  \caption{\parbox[t]{.8\linewidth}{Learning curves for path 2.  Our method (hierarchical latent) reuses the low-level policy learned for path 1.}}
  \label{fig:learning_curve_c}
\end{subfigure}
\caption{Learning curves of a flat policy, a hierarchical policy with latent commands and an expert hierarchical policy. We plot the average of 5 statistical runs with shaded area representing the standard error.}
\label{fig:learning_curve}
\end{figure}

In Fig.~\ref{fig:learning_curve} we show learning curves for $3$ policies, a flat policy, hierarchical policy with expert-designed, pre-trained low-level, and a hierarchical policy with latent command space (our method). The policies are trained on $2$ different paths. All three methods succeed in solving the task of following the first path (Fig.~\ref{fig:learning_curve_b}). For the second path, our method is able to solve the task significantly faster than other policies (Fig.~\ref{fig:learning_curve_c}). On the second path, the flat policy has to learn the parameters from scratch. The expert policy's high-level learns to use the same low-level policy used in the first path. This low-level policy was pre-trained (see Appendix). Therefore, the expert policy needs extra training time to learn both levels separately. On the other hand, both  levels of our latent command based hierarchical policy are trained from scratch on the first path. The best performing policy uses a $4$ dimensional latent space. We can see that this policy can still reuse the same low-level and $4$D latent commands to adapt quickly to a new task.

Fig.~\ref{fig:sim_paths} shows how the robot trained with a hierarchical policy behaves in simulation. It successfully follows the path using steering behaviors. Complete trajectories can be seen in Fig.~\ref{fig:sim_paths_b}. Markers along the trajectory show points at which the high-level becomes active and computes the next latent command and duration. The low-level policy was only trained on the first path and is reused for the second path.

\begin{figure*}
\centering
\begin{subfigure}{.65\columnwidth}
  \centering
  \includegraphics[width=\linewidth]{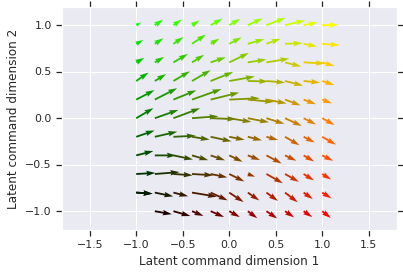}
  \caption{\parbox[t]{.7\linewidth}{Low-level behaviors  sampled from  a 2D  latent command space. Vector directions correspond to the movement direction of the robot. Vector length is proportional to the distance covered.}}
  \label{fig:latentspaceanalysis_a}
\end{subfigure}%
\begin{subfigure}{.65\columnwidth}
  \centering
  \includegraphics[width=\linewidth]{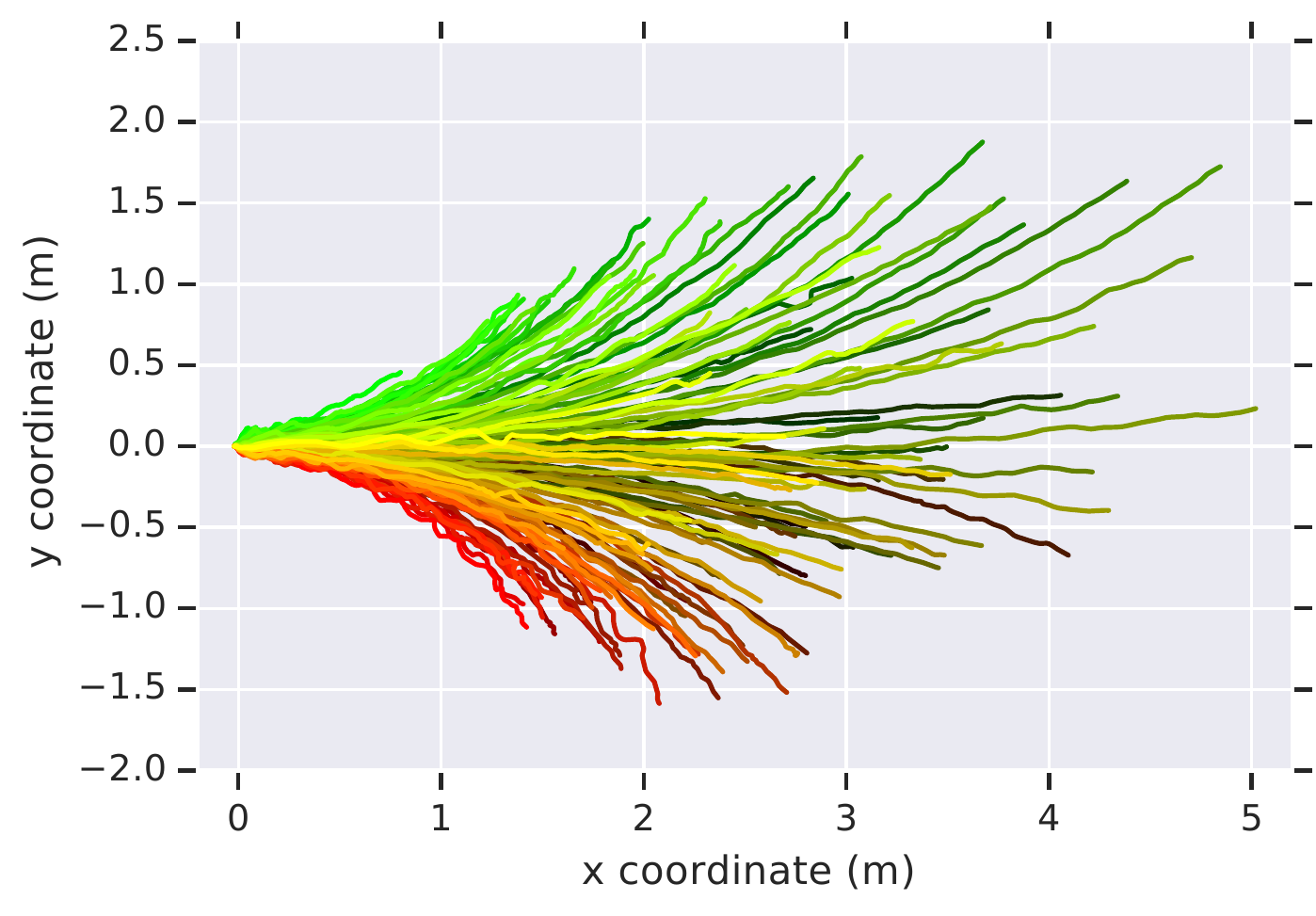}
  \caption{\parbox[t]{.7\linewidth}{Low-level behaviors for different latent commands (colors correspond to Fig.~\ref{fig:latentspaceanalysis_a}). Notice that while diverse, the low-level behaviors are biased towards left turns because of the task at-hand.}}
  \label{fig:latentspaceanalysis_b}
\end{subfigure}%
\begin{subfigure}{.65\columnwidth}
  \centering
  \includegraphics[width=\linewidth]{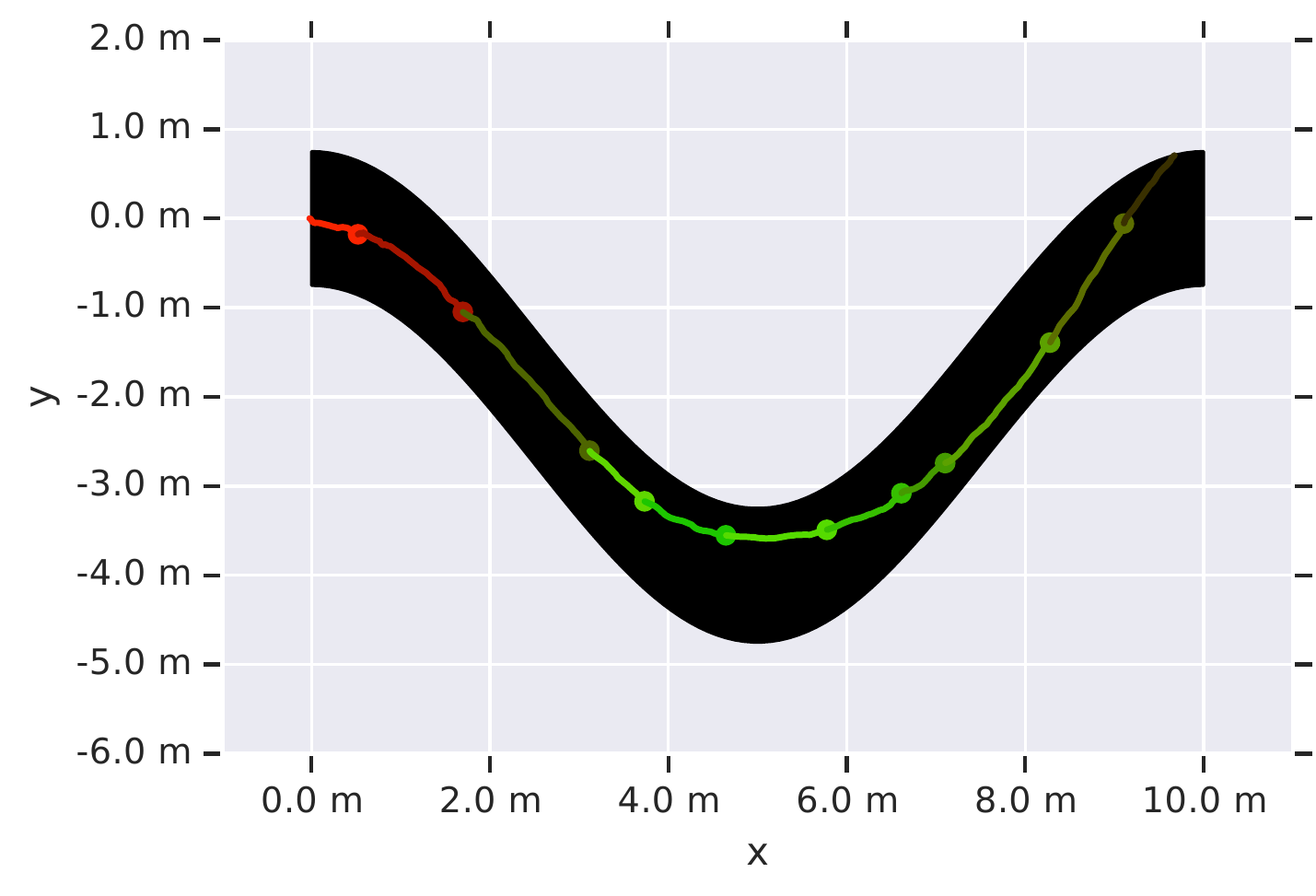}
  \includegraphics[width=\linewidth]{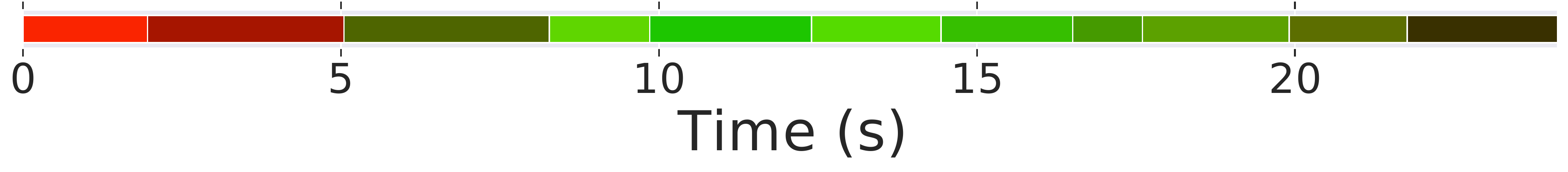}
  \caption{\parbox[t]{.7\linewidth}{Sample trajectory of the HRL policy with a 2D latent command space. Dots indicate new high-level commands. The timeline shows the high-level activations.}}
  \label{fig:latentspaceanalysis_c}
\end{subfigure}
\caption{Analysis of latent command space $\bm{l}$ and  low-level duration $d$.}
\label{fig:latentspaceanalysis}
\end{figure*}

To simplify the analysis, we study a $2$ dimensional latent command space learned by our method in Fig.~\ref{fig:latentspaceanalysis}. We evaluated the low-level for different points in the latent space. In Fig.~\ref{fig:latentspaceanalysis_a} we show the movement direction of the robot when giving different points in latent space as commands to the low-level and executing the low-level for a fixed number of steps ($1000$). The length of the arrow is proportional to the distance covered. Corresponding color-coded robot trajectories are shown in Fig.~\ref{fig:latentspaceanalysis_b}. We can observe that for the path following task, robot steering behaviors of varying velocities emerge automatically as low-level behaviors. The high-level uses these steering behaviors to navigate different parts of the path as show in Fig.~\ref{fig:latentspaceanalysis_b}. Moreover, the high-level also decides a variable duration for each latent command (see Fig.~\ref{fig:latentspaceanalysis_b}). We can observe that for straighter parts of the path, the high-level selects a longer duration to go forward, while for curved parts, it  switches latent commands more frequently.

\begin{figure}[htpb]
\centering
\includegraphics[width=.5\linewidth]{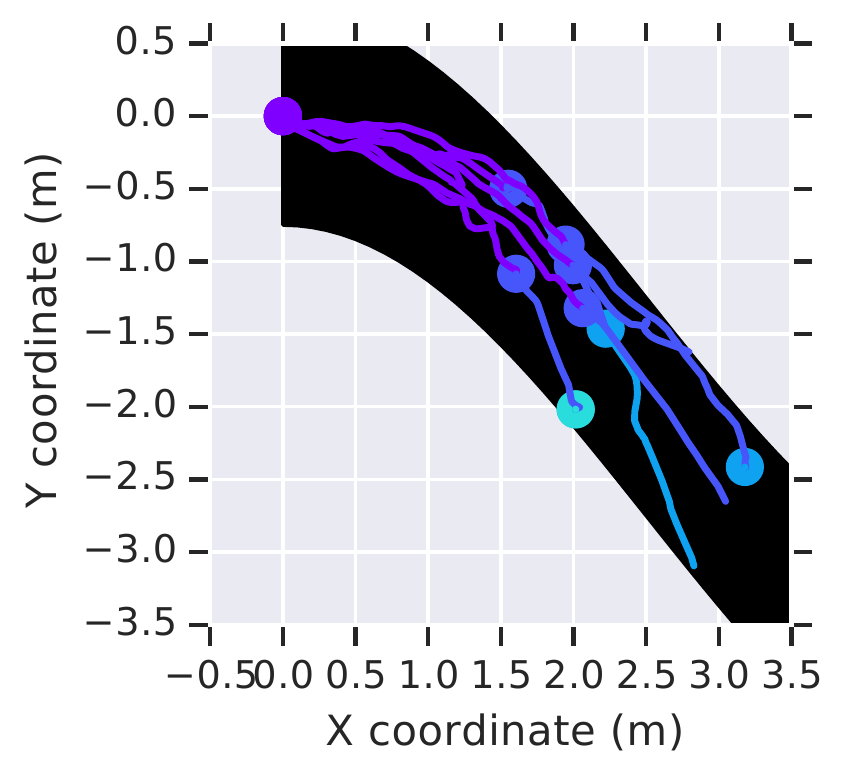}%
\includegraphics[width=.5\linewidth]{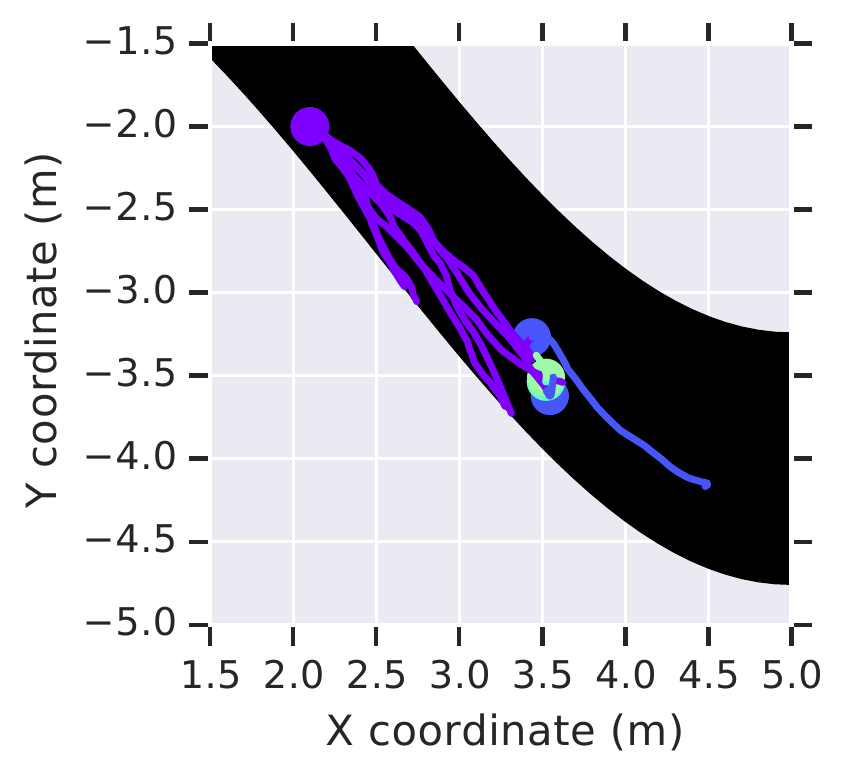}
\caption{The trajectories of the real robot measured with motion capture while using a trained HRL policy at different segments of the path.}
\label{fig:robottrajectory}
\end{figure}

\subsection{Hardware Validation}
Finally, we validate our results by transferring an  HRL policy to a real robot and recording the resulting trajectories. We use a motion capture system (PhaseSpace Impulse X2E) to estimate the robot's current position and heading, which is then fed into the high-level policy. Since our architecture allows execution of the different levels at different frequencies, it is sufficient to transmit motion capture data to the high-level policy at a much lower rate compared to low-level sensor data such as IMU readings.

Because of the limited capture volume in our lab setting, we were only able to track the robot's trajectory along part of the task (see Fig.~\ref{fig:intro-path} and~\ref{fig:robottrajectory}). To overcome this limitation, we recorded shorter robot trajectories starting at the origin. We then virtually moved the robot down the path by adding an offset to the motion capture's position estimate and recorded another set of trajectories. Note the significant variance for the real trajectories at the start of the path due to slippage of the legs during dynamic turning gaits.





\section{CONCLUSION}
We presented a hierarchical control approach particularly suited for legged robots. 
By separating the architecture into two parts, a high-level and a low-level policy network, and jointly training them, we obtained a number of advantages over previous algorithms.

First, the architecture is agnostic to the task: we do not need to manually pick or pretrain the behaviors (primitives) of the low-level policy. As a consequence we also remove the need to design individual reward functions for each behavior.
In fact, our algorithm outperforms a similar setup in which the low-level behaviors are predefined.

Secondly, our method can be used to bootstrap when training on a new task by transferring the trained low-level policy.

Finally, the high-level and low-level policies operate at different timescales and can use different state representations.
This is of particular practical importance, since motor commands should be able to be calculated in mere milliseconds by a low-level policy for safety and stability reasons. High-level signals such as rewards or position estimates are often updated at much lower frequencies and might have to be transmitted via a wireless connection. 
Our approach provides a natural way to decouple these timescales.

The task at hand allowed us to study the results in detail in both simulation and hardware to validate our approach and implementation. We show that given the path following task, the steering behaviors automatically emerge in a latent space, and the robot can easily adapt to a new path with low-level transfer. We also deployed these policies to  hardware to validate the learned hierarchical policy.

In future work, we plan to apply this algorithm on tasks requiring a high level of agility in more complex environments. As an example, if the robot has to jump over an obstacle or climb stairs, manually defining a set of low-level behaviors will become even more cumbersome. We believe that the latent command space will allow us to tackle these challenges through automatic discovery of the complex primitives required to solve the task. 
In addition, we are planning to incorporate more complex  sensors such as camera images, which naturally operate at different timescales and require significant computational power. In this case our approach would allow for distributed  processing, without compromising performance.

\section*{APPENDIX}
As part of the baselines, a low-level expert steering policy is trained separately. This policy is controlled by a scalar input from the high-level $l$, which determines the target direction. We train the policy using the ARS algorithm by rewarding the magnitude of the average steering angle over the past $50$ timesteps. The reward is capped by the input $l$. Then another  component (weighted by $\alpha$) is added to the reward for moving forward, which is capped by a \mbox{fixed value, $r^{\mbox{fw}}_{\mbox{cap}}$}:
\begin{align}
    r^{\mbox{steer}}(t) &= \min\left(l, \theta_t^{\mbox{steer}}\right) \\
    r^{\mbox{fw}}(t) &= \min\left(r^{\mbox{fw}}_{\mbox{cap}}, \bm{x}(t)-\bm{x}(t-1)\right) \\
    r(t) &= r^{\mbox{steer}}(t) + \alpha r^{\mbox{fw}}(t) \\
    R &= \sum_{t \ge 1}r(t).
\end{align}

\begin{figure}
  \centering
  \includegraphics[width=.8\linewidth]{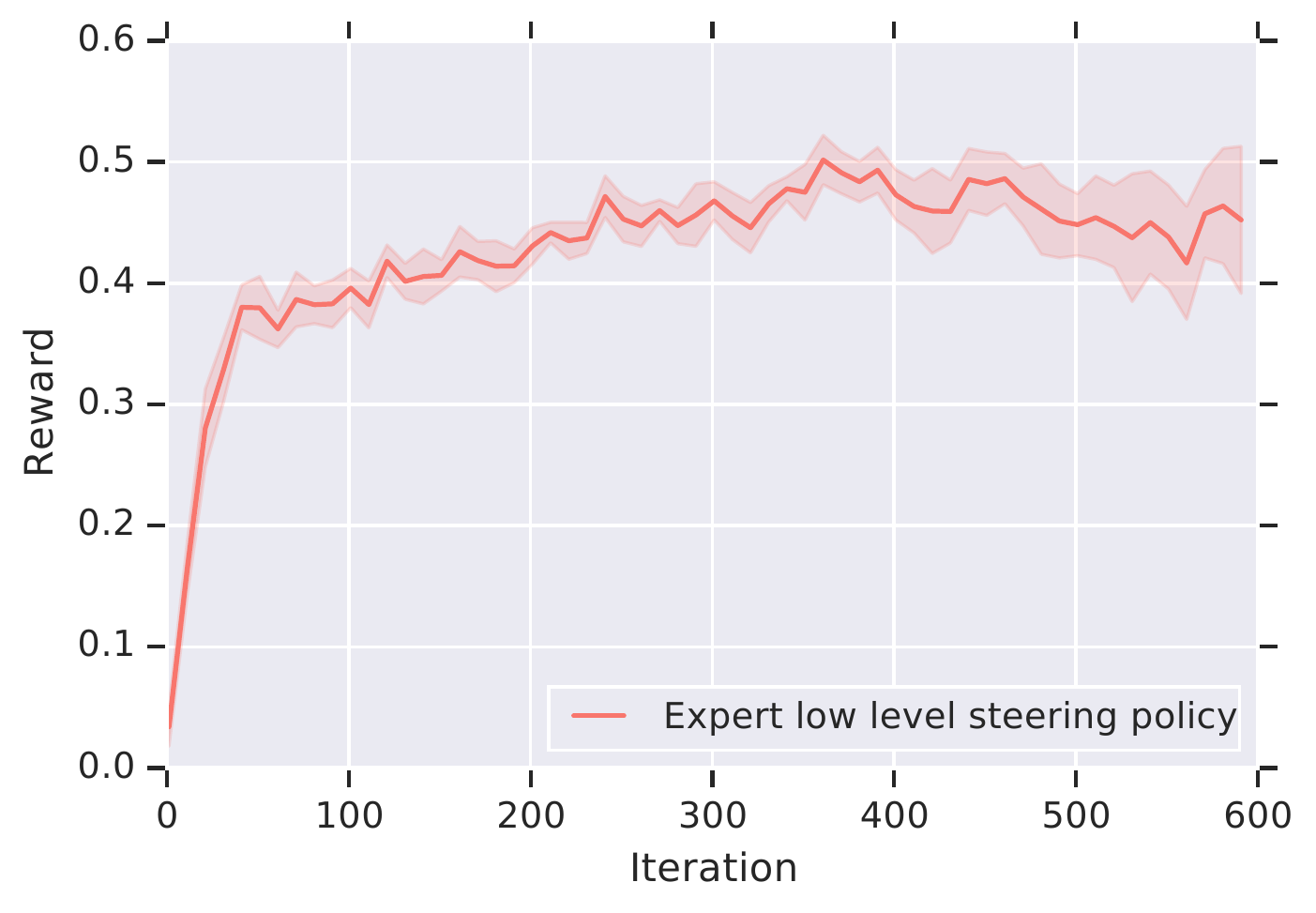}
  \caption{\parbox[t]{.8\linewidth}{Learning curve for the pre-training phase of the expert low-level policy.}}
  \label{fig:expertll_b}
\end{figure}

\begin{figure}
  \centering
  \includegraphics[width=.8\linewidth]{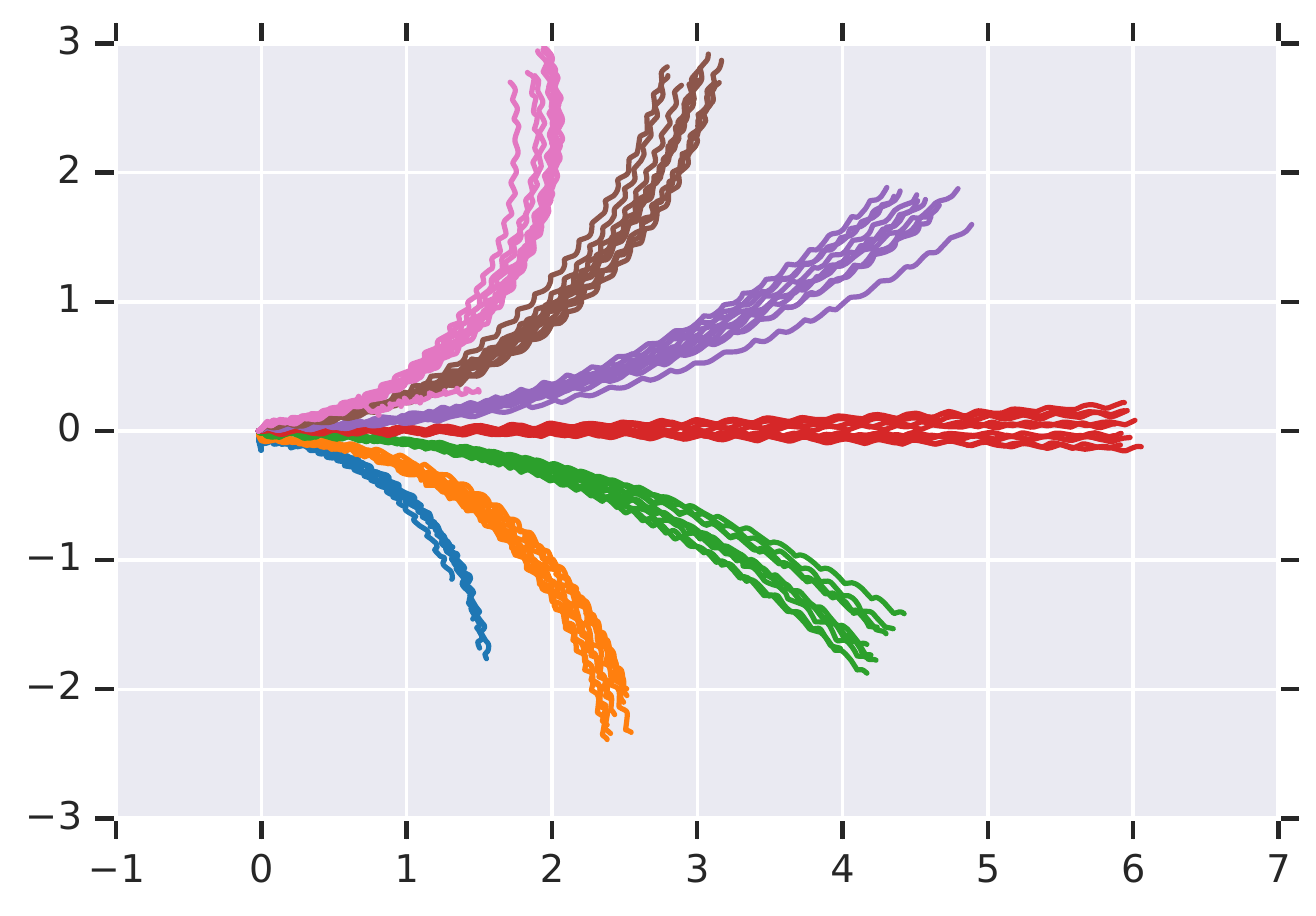}
  \caption{\parbox[t]{.8\linewidth}{Expert low-level policy with different inputs (axes in $\si{\meter}$).}}
  \label{fig:expertll_a}
\end{figure}

For training, we randomly sample an input $l$ from a uniform distribution for each episode. The learning curve for training this policy is shown in Fig.~\ref{fig:expertll_b}. Sample trajectories after training are shown in Fig.~\ref{fig:expertll_a}.


\section*{ACKNOWLEDGMENT}
We would like to thank Jie Tan, Tingnan Zhang, Erwin Coumans, Sehoon Ha (Robotics at Google), Honglak Lee, Ofir Nachum (Google Brain), and Arun Ahuja (DeepMind) for insightful discussions.


\balance
\bibliographystyle{unsrt}
\bibliography{sample}

\end{document}